\documentclass[letterpaper, 10 pt, conference]{ieeeconf}
\IEEEoverridecommandlockouts
\let\labelindent\relax

\usepackage{handy_style}
\usepackage{booktabs}
\usepackage{threeparttable}
\usepackage{bm}
\let\oldnl\nl
\newcommand{\nonl}{\renewcommand{\nl}{\let\nl\oldnl}}
\newcolumntype{x}{>{\columncolor{MistyRose}}c}
\newcolumntype{y}{>{\columncolor{LightCyan1}}c}

\newcommand\blfootnote[1]{%
  \begingroup
  \renewcommand\thefootnote{}\footnote{#1}%
  \addtocounter{footnote}{-1}%
  \endgroup
}

\acrodef{urdf}[URDF]{Unified Robot Description Format}

\title{\vspace*{-16pt}\LARGE \bf Benchmarking Bimanual Occluded Grasping through State-Grounded Progress and Interaction Precision}%

\author{Haodi Liu\textsuperscript{1}, Xinhang Yang\textsuperscript{1}, Zijie Wei\textsuperscript{2},
  Sen Cui\textsuperscript{1}, Zeyu Zhang\textsuperscript{3}, and Changshui Zhang\textsuperscript{1}}

\newcommand{\authoraffiliations}{%
  \textsuperscript{1}Beijing National Research Center for Information Science and Technology (BNRist),
  Department of Automation, Tsinghua University, Beijing, China.\par
  \textsuperscript{2}The High School Affiliated with Renmin University of China, Beijing, China.\par
  \textsuperscript{3}State Key Laboratory of General Artificial Intelligence,
  Beijing Institute for General Artificial Intelligence (BIGAI), Beijing, China.}

\hypersetup{
  pdftitle={Benchmarking Bimanual Occluded Grasping through State-Grounded Progress and Interaction Precision},
  pdfauthor={Haodi Liu, Xinhang Yang, Zijie Wei, Sen Cui, Zeyu Zhang, Changshui Zhang}
}
\begin{document}

\twocolumn[{
\renewcommand\twocolumn[1][]{#1}
\maketitle
\thispagestyle{empty}
\pagestyle{empty}
\vspace{-0.5cm}
\begin{center}
    \centering
    \captionsetup{type=figure}
       \includegraphics[width=\linewidth]{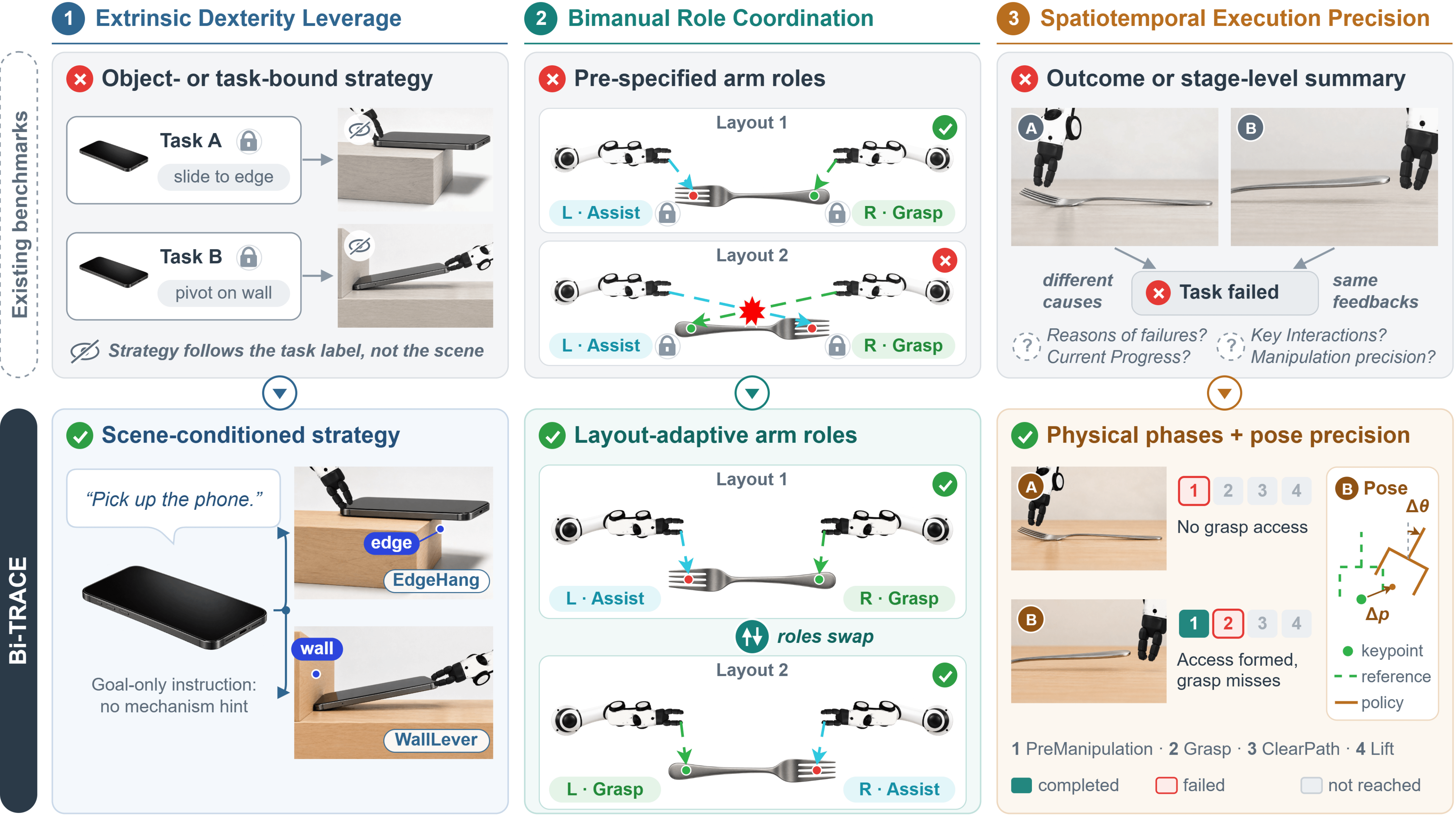}
       \caption{\textbf{Existing benchmarks vs.\ Bi-TRACE on three core capabilities of bimanual occluded grasping.} Each column contrasts a limitation of existing benchmarks (top) with Bi-TRACE (bottom): \ding{182} object- or task-bound strategies versus scene-conditioned strategy selection for the same object under a goal-only instruction; \ding{183} pre-specified arm roles versus layout-adaptive role swapping; \ding{184} outcome- or stage-level summaries versus state-grounded phase tracking, which separates failing to create grasp access (A) from missing the grasp (B), and keypoint-relative pose errors at critical interactions.}
       \label{fig:intro_bi}
\end{center}
}]
\blfootnote{\authoraffiliations}
\blfootnote{This work has been submitted to the IEEE for possible publication. Copyright may be transferred without notice, after which this version may no longer be accessible.}

\begin{abstract}
In bimanual occluded grasping (BOG), one arm creates and, when needed, maintains space for the other to grasp and lift an otherwise inaccessible object. BOG is a representative setting for complex manipulation, combining precise contact, coordination between arms, and multi-stage physical dependencies. As task success alone offers little insight into the execution process, we present Bi-TRACE, a diagnostic benchmark that systematically evaluates bimanual policies along four complementary dimensions: selecting appropriate grasp-enabling strategies from visual observations, assigning assisting and grasping roles according to scene layout, achieving verifiable physical progress across successive task stages, and accurately posing grippers at critical interaction sites. We train six general-purpose policies jointly on four tasks and test them on new layouts of these tasks and on four additional tasks with unseen objects and recolored surroundings. Policies often attempt suitable actions without producing the required physical changes: 98\% of failures on the training tasks occur before a firm grasp is secured, and the best policy succeeds in only 30.8\% of episodes on these tasks and 2.5\% on the new tasks. Moreover, accurate gripper positioning alone does not guarantee successful task execution. These systematic evaluations reveal capability boundaries of current policies and identify concrete targets for improving bimanual grasping and multi-stage manipulation.
\end{abstract}

\section{Introduction}

Dexterous manipulation demands more than reaching for accessible objects: when surrounding structures occlude a grasp, a robot must first reshape object--environment contacts to make grasping feasible~\cite{zhou2023learning,zhang2023reinforcement,wu2023learning,kobashi2025leveraging}.
Bimanual occluded grasping (BOG) brings together challenges common to complex manipulation---multi-stage physical dependencies, precise spatial contact, and tight temporal coordination---within a fundamental grasping task. Here, one arm can create and, when necessary, maintain grasp access while the other secures the object for lifting~\cite{grannen2023stabilize,liu2025voxact,yamada2025combo}.
This makes BOG a representative setting for examining whether general-purpose policies can learn from task demonstrations to produce physical effects that support subsequent actions, rather than merely reproduce plausible motions. A plausible push or press may fail to create grasp access, or the created access may be lost during grasping. Task success alone cannot distinguish these failures (\cref{fig:intro_bi}, top right). We therefore present \textbf{Bi-TRACE}, a benchmark that diagnoses this causal chain by tracking the variations of physical states and evaluating the precision of critical interactions (\cref{fig:intro_bi}, bottom right).

Systematically evaluating this causal chain requires decomposing BOG execution into three measurable aspects. First, \textbf{Extrinsic Dexterity Leverage} (\cref{fig:intro_bi}\,\ding{182}) demands that the assisting arm actively exploit environmental structures (e.g., table edges, walls) and object geometry to transition occluded objects into graspable states, ensuring pre-grasp contacts yield desired physical outcomes rather than mere motion replay. Second, the random nature of object poses and affordance locations necessitates adaptive \textbf{Bimanual Role Coordination} (\cref{fig:intro_bi}\,\ding{183}): policies must adaptively allocate assisting and grasping roles based on kinematic reachability and motion cost, as fixed role assignments can lead to trajectory collisions and inter-arm interference. Finally, translating an exposed grasp into a stable lift requires \textbf{Spatiotemporal Execution Precision} (\cref{fig:intro_bi}\,\ding{184}), where policies must execute phase-appropriate actions precisely at the target locations.

Despite recent expansions in task diversity~\cite{grotz2025twin,mu2025robotwin,chen2025robotwin,chen2025robotwinchallenge,chernyadev2025bigym,shen2026bipremanip,peng2026bicoord}, existing bimanual benchmarks present three major limitations that hinder systematic evaluation of the core capabilities along the causal chain (\cref{fig:intro_bi}, top). First, while prior benchmarks include pre-grasp actions~\cite{grotz2025twin,shen2026bipremanip,maurya2026imbench}, scene-conditioned mechanism selection remains underexplored: whether policies can choose appropriate grasp-enabling strategies as object geometry and available environmental structures vary, rather than relying on object- or task-specific strategy associations.
Second, most benchmarks pre-specify arm roles, which can induce kinematic conflicts under randomized layouts, restrict demonstrations to fixed-role scenarios, and preclude evaluation of adaptive coordination via role swapping.
Third, recent evaluations assess execution quality and stage-wise progress~\cite{shen2026bipremanip,peng2026bicoord,wang2026roboeval,julg2026duobench,yang2026bench2dex}, but provide limited insight into how successive actions establish and preserve the physical conditions needed for later stages, and how accurately grippers are posed during critical contact and grasping interactions.

To address these limitations, Bi-TRACE evaluates the three core capabilities throughout BOG execution (\cref{fig:intro_bi}, bottom). First, Bi-TRACE evaluates scene-conditioned selection of grasp-enabling strategies leveraging support edges, vertical walls, and object geometry. Different combinations of object geometry and environmental structures can require different mechanisms, which policies must infer from visual observations rather than task labels; language instructions, when used, provide no mechanism hints.
Second, Bi-TRACE assesses adaptive bimanual coordination by allowing either arm to serve as the assisting or grasping arm, enabling dynamic role swapping based on kinematic feasibility, which reduces interference from fixed role assignments and expands demonstration coverage to layouts that fixed-role approaches cannot handle.
Third, for fine-grained spatiotemporal diagnosis, Bi-TRACE tracks how physical conditions are created, maintained, and exploited as execution progresses. It further evaluates the spatial precision of critical contact and grasping actions, revealing not only where execution fails but also how accurately policies perform the interactions needed to achieve the intended physical effects.

In summary, our contribution is threefold:
\begin{enumerate}[wide=\parindent]
    \item We frame BOG as a representative setting where multi-stage physical dependencies, precise contact, and coordinated execution converge. This motivates evaluating whether policies actually create, maintain when needed, and exploit the physical conditions for subsequent actions, rather than merely reproduce plausible motions.

    \item We present Bi-TRACE, a benchmark built on three grasp-enabling mechanisms, adaptive arm-role assignment, state-grounded phase tracking, and functional keypoints. Its complementary full-task and key-interaction evaluations localize phase-wise failures and quantify the keypoint-relative precision of critical interactions, exposing execution bottlenecks that binary task success obscures.

    \item We benchmark six general-purpose policies on held-in and held-out tasks, revealing recurring failures in access creation and coordinated grasp formation, often despite mechanism-consistent behavior. The results further distinguish local pose accuracy from autonomous physical effectiveness, providing testable targets for improving contact execution, access maintenance, and bimanual coordination.
\end{enumerate}

\section{Related work}

\subsection{Bimanual Occluded Grasping}

Bimanual occluded grasping (BOG) extends occluded grasping, where supporting surfaces or surrounding structures block desired grasps, by requiring dual-arm systems to actively establish graspable configurations. Related work spans two streams. The first studies extrinsic dexterity for ungraspable objects~\cite{zhou2023learning,zhang2023reinforcement,wu2023learning,kobashi2025leveraging}, using non-prehensile pre-grasp actions, such as pushing~\cite{zhang2023reinforcement}, sliding~\cite{wu2023learning}, or wall pivoting~\cite{kobashi2025leveraging}, to enable grasping. However, these methods are mostly single-arm and rely on fixed structures or predefined interaction resources. The second explores bimanual manipulation~\cite{grannen2023stabilize,liu2025voxact,yamada2025combo}, using a secondary arm to stabilize objects~\cite{grannen2023stabilize,liu2025voxact} or constrain object motion to support grasping~\cite{yamada2025combo}. These methods demonstrate how physical constraints can enable grasping; Bi-TRACE turns this principle into a diagnostic benchmark for evaluating whether policies can create, maintain, and exploit grasp access through coordinated interactions.

\subsection{Bimanual Manipulation Benchmarks}
Recent bimanual benchmarks have advanced task diversity, coordinated manipulation, and diagnostic evaluation. TWIN~\cite{grotz2025twin}, RoboTwin~\cite{mu2025robotwin,chen2025robotwin,chen2025robotwinchallenge}, and BiGym~\cite{chernyadev2025bigym} broaden task coverage and dual-arm evaluation across simulated and real-world settings. Beyond task diversity, BiPreManip~\cite{shen2026bipremanip} explores affordance-guided preparatory actions that facilitate subsequent manipulation, while BiCoord~\cite{peng2026bicoord} examines long-horizon coordination through stage-annotated tasks. IMBench~\cite{maurya2026imbench} investigates physical reasoning and execution, including grasp-enabling actions that make obstructed objects accessible. Recent diagnostic benchmarks further move beyond task success: RoboEval~\cite{wang2026roboeval} assesses execution quality, DuoBench~\cite{julg2026duobench} analyzes stage-wise coordination and progress, and Bench2Dex~\cite{yang2026bench2dex} evaluates stable success and physical progress across dexterous hands. Bi-TRACE instead evaluates how policies create, maintain, and exploit grasp access through precise bimanual coordination. By jointly assessing physical progress and interaction accuracy, it reveals execution bottlenecks that full-task evaluation may overlook.

\section{Method}

\begin{figure*}[!t]
    \centering
    \includegraphics[width=\textwidth]{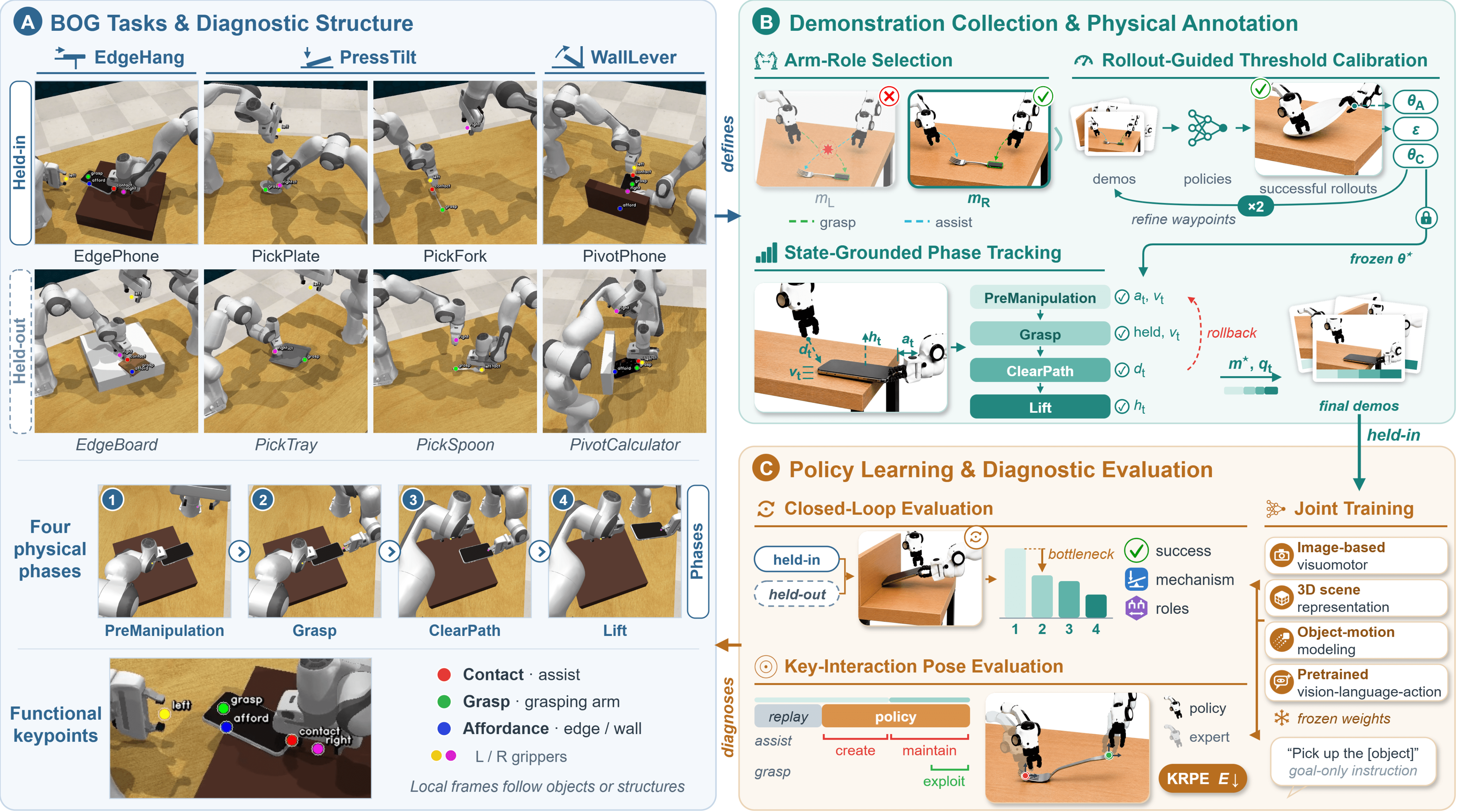}
    \caption{\textbf{Overview of Bi-TRACE.} (A) Four held-in tasks and four held-out tasks with unseen objects cover three grasp-enabling mechanisms, four physical phases, and functional keypoints. (B) Adaptive arm-role selection, rollout-guided threshold calibration, and state-grounded phase tracking yield demonstrations annotated with arm roles and physical progress. (C) Policies trained jointly on held-in demonstrations are evaluated through closed-loop rollouts on both task sets for phase progress, and through rollouts starting from expert-replayed near-contact states for keypoint-relative pose accuracy.
    }
    \label{fig:method_overview}
\end{figure*}

Bi-TRACE evaluates BOG by tracking the physical conditions that enable successive actions and measuring the precision of critical interactions (\cref{fig:method_overview}). We first define the task suite with shared physical phases and functional keypoints (\cref{sec:bog_tasks}), then collect and annotate demonstrations using adaptive arm-role selection and rollout-calibrated physical criteria (\cref{sec:demonstration_collection}). These demonstrations support joint policy training and complementary evaluations of full-task execution and key-interaction accuracy (\cref{sec:policy_learning_evaluation}).

\subsection{BOG Tasks \& Diagnostic Structure}
\label{sec:bog_tasks}

The task suite covers three grasp-enabling mechanisms (\cref{fig:method_overview}A, top): EdgeHang, PressTilt, and WallLever exploit support edges, object geometry, and wall constraints, respectively.
Across them, the four held-in tasks progressively combine execution demands: \textit{EdgePhone} emphasizes creating grasp access, as its overhang is largely self-maintaining; \textit{PickPlate} adds sustained assisting-arm pressure to maintain tilt during grasping; \textit{PickFork} applies PressTilt to a slender object whose narrow contact and grasp regions demand spatial precision; and \textit{PivotPhone} combines precise contact, pose maintenance, and coordinated grasping under wall constraints. Policies must therefore infer grasp-enabling strategies from object geometry and scene constraints.

The held-in tasks provide training demonstrations and independent test layouts, which vary mainly in the poses of the target and its environmental structures relative to the two arms.
The four held-out tasks retain the same mechanisms but use new objects of similar shape and manipulation characteristics, along with recolored surroundings. Policies trained only on held-in demonstrations are evaluated on them to test adaptation to new objects and appearance.

All eight tasks share four physical phases (\cref{fig:method_overview}A, middle): \textbf{PreManipulation} establishes sufficient, stable grasp access; \textbf{Grasp} uses this access to establish a firm grasp; \textbf{ClearPath} requires the assisting gripper to be empty and retracted from the object; and \textbf{Lift} raises the held object to a preset height. These phases track progress through the physical conditions required by subsequent actions. Three types of functional keypoints with local coordinate frames specify critical interaction poses (\cref{fig:method_overview}A, bottom): \textbf{Contact} marks the assisting arm's pre-manipulation site, \textbf{Grasp} the grasping arm's grasp site, and \textbf{Affordance} the support edge or wall exploited by EdgeHang and WallLever; PressTilt relies on object geometry instead.

\subsection{Demonstration Collection \& Physical Annotation}
\label{sec:demonstration_collection}

\looseness=-1 For each initialized scene, we configure two mirrored waypoint schemes that swap the arm roles: left grasp/right assist and right grasp/left assist (\cref{fig:method_overview}B, upper left).
The assisting arm creates and, when needed, maintains grasp access through pre-manipulation, while the grasping arm secures and lifts the object.
\textbf{Arm-Role Selection} chooses between the schemes based on reachability and motion cost in the current layout (Algorithm~\ref{alg:demo_collection}, lines~\ref{ln:role_begin}--\ref{ln:role_end}).
As the assisting arm's approach initiates pre-manipulation, on which all subsequent actions depend, we check its collision-free reachability to the first waypoint and weight its motion cost more heavily.

\begin{table}[t]
\centering
\caption{Physical quantities and thresholds for state-grounded phase tracking (EH: EdgeHang; PT: PressTilt; WL: WallLever).}
\label{tab:physical_conditions}
\footnotesize
\setlength{\tabcolsep}{3pt}
\begin{tabularx}{\linewidth}{@{}
  l
  >{\raggedright\arraybackslash}X
  >{\raggedright\arraybackslash}p{0.37\linewidth}@{}}
\toprule
\textbf{Category} & \textbf{Physical quantity} & \textbf{Threshold and requirement} \\
\midrule
Access &
$a_t$: overhang for EH; height of Grasp keypoint for WL and PT. &
$a_t>\theta_A$ for EH; $a_t\geq\theta_A$ for WL and PT. \\
\addlinespace
Stability &
$\nu_t$: combined linear and angular speed of the object. &
$\nu_t<\varepsilon$ over consecutive timesteps. \\
\addlinespace
Clearance &
$d_t$: distance from the assisting end effector to the target object. &
$d_t\geq\theta_C$. \\
\addlinespace
Elevation &
$h_t$: world height of the target object origin. &
$h_t\geq\theta_E$. \\
\bottomrule
\end{tabularx}
\end{table}

\begin{algorithm}[t]
\SetAlCapFnt{\footnotesize}\SetAlCapNameFnt{\footnotesize}
\caption{Demonstration Collection \& Physical Annotation}
\label{alg:demo_collection}
\footnotesize
\SetKwProg{Fn}{Function}{:}{}
\SetFuncSty{textrm}
\SetKwFunction{FSelect}{SelectRoles}
\SetKwFunction{FTrack}{Track}
\KwIn{Sampled layouts of a held-in task; mirrored waypoint schemes $M=\{m_L,m_R\}$ for left-/right-arm grasping; assisting-arm weight $\beta>1$; lifting threshold $\theta_E$.}
\KwOut{Thresholds $\theta_A,\varepsilon,\theta_C$; final demonstrations $\mathcal{D}$.}

\tcp{Arm-Role Selection}
\Fn{\FSelect{$s_0$}}{\nllabel{ln:role_begin}
    \lForEach{$m\in M$}{$f_m\gets\operatorname{AssistReachable}(s_0,m)$}
    \lIf{$f_{m_L}\neq f_{m_R}$}{\KwRet{the feasible scheme}}
    \KwRet{$\arg\min_{m\in M}\bigl[c_g(m)+\beta\,c_a(m)\bigr]$}\;\nllabel{ln:role_end}
}
\tcp{Rollout-Guided Threshold Calibration}
\For{round $=1,2$}{\nllabel{ln:cal_begin}
    $\mathcal{D}\gets$ execute scheme \FSelect{$s_0$} for each sampled $s_0$\;
    $\mathcal{R}\gets$ verified successful rollouts of policies trained on $\mathcal{D}$\;
    $(\theta_A,\varepsilon,\theta_C)\gets$ means of $(a_t,\nu_t,d_t)$ at transition events in $\mathcal{R}$\;
    refine waypoints in $M$ to satisfy $\theta_A,\varepsilon,\theta_C$\;\nllabel{ln:cal_end}
}
\tcp{State-Grounded Phase Tracking}
\Fn{\FTrack{$t,q,b$}}{\nllabel{ln:track_begin}
    \lIf{$q=5$}{\KwRet{$(q,b)$}}
    $\mathit{space}\gets a_t\geq\theta_A$;\ \ $\mathit{held}\gets$ grasping arm of $m^\star$ holds target\;
    $\mathit{clear}\gets$ assisting gripper empty $\land\,d_t\geq\theta_C$;\ \ $\mathit{lift}\gets h_t\geq\theta_E$\;
    $\mathit{access}\gets S(\mathit{space})$;\ \ $\mathit{firm}\gets S(\mathit{held})$\;
    $\mathit{maintain}\gets(\mathrm{true},\mathit{access},\mathit{held},\mathit{held})$\;
    $\mathit{progress}\gets(\mathit{access},\mathit{firm},\mathit{clear},\mathit{lift})$\;
    \tcp{Roll back, then advance.}
    \lIf{$\neg\mathit{maintain}[q]$}{$q\gets1$}
    \lWhile{$q<5\land\mathit{maintain}[q]\land\mathit{progress}[q]$}{$q\gets q+1$}
    \KwRet{$\bigl(q,\max(b,q-1)\bigr)$}\;\nllabel{ln:track_end}
}
\tcp{Final demonstration collection}
$\mathcal{D}\gets\emptyset$\;\nllabel{ln:final_begin}
\ForEach{sampled layout $s_0$}{
    $m^\star\gets$ \FSelect{$s_0$};\ \ $(q,b)\gets(1,0)$\;
    \ForEach{step $t$ while executing $m^\star$}{
        record observations, actions, and physical states\;
        $(q,b)\gets$ \FTrack{$t,q,b$};\ \ label step $t$ with $(m^\star,q)$\;
    }
    \lIf{$b=4$}{add the episode to $\mathcal{D}$}\nllabel{ln:final_end}
}
\KwRet{$\theta_A,\varepsilon,\theta_C,\mathcal{D}$}\;

{\let\nl\relax
\vspace{2pt}
\noindent\hspace*{-1em}%
\begin{minipage}{\dimexpr\linewidth-2em-2pt\relax}
\footnotesize
The grasping and assisting costs $c_g$ and $c_a$ sum position and orientation differences from each arm's initial gripper pose to its waypoints, with greater weight on the first waypoint.
$S(x)$ requires $x$ and $\nu_t<\varepsilon$ over consecutive timesteps.
Condition vectors are indexed by phases 1--4; $b$ records the highest completed phase and never decreases.
\end{minipage}\par
}
\end{algorithm}

The geometric and stability conditions of each phase are evaluated by comparing physical measurements with the thresholds in \autoref{tab:physical_conditions}, which we determine through \textbf{Rollout-Guided Threshold Calibration} (\cref{fig:method_overview}B, upper right).
Calibration serves two linked purposes: establishing reasonable subgoal criteria for task execution and aligning expert demonstrations with feasible motion patterns of learned policies.
The lifting threshold $\theta_E$ follows the task goal, while $\theta_A$, $\varepsilon$, and $\theta_C$ are calibrated task-wise over two rounds (lines~\ref{ln:cal_begin}--\ref{ln:cal_end}).
Starting from expert waypoints tuned for reliable completion across sampled layouts, each round collects demonstrations, trains imitation policies, and executes them on the same layouts, where successful rollouts are more readily available.
We manually verify successful rollouts, mark the physical events corresponding to phase transitions, and average the synchronized measurements at these events.
The resulting thresholds reflect representative, physically effective states in successful rollouts, and the expert waypoints are refined to satisfy them.

\looseness=-1 With arm roles and transition thresholds determined, \textbf{State-Grounded Phase Tracking} updates the current phase $q$ from the evolving physical state, identifying the subgoal that current actions should achieve (\cref{fig:method_overview}B, lower left; lines~\ref{ln:track_begin}--\ref{ln:track_end}).
Here, $\mathit{space}$ indicates sufficient geometric grasp access ($a_t$ reaching $\theta_A$), and $\mathit{held}$ indicates that the grasping arm holds the target.
The shared stability check $S(x)$ requires condition $x$ and low object speed ($\nu_t<\varepsilon$) to hold over consecutive timesteps.
Thus, $\mathit{access}=S(\mathit{space})$ and $\mathit{firm}=S(\mathit{held})$ indicate stable grasp access and a firm grasp, respectively.
The condition $\mathit{clear}$ requires the assisting gripper to be empty and $d_t\geq\theta_C$, while $\mathit{lift}$ requires $h_t\geq\theta_E$.
For each phase, the \emph{maintenance condition} specifies what must remain true to preserve prior progress, while the \emph{progression condition} confirms completion of the current subgoal; advancement requires both.
PreManipulation has no maintenance requirement and advances when $\mathit{access}$ holds.
Grasp requires continued $\mathit{access}$ and advances when $\mathit{firm}$ holds.
After grasping, ClearPath and Lift require only $\mathit{held}$ for maintenance, allowing the object to move with the gripper; they advance when $\mathit{clear}$ and $\mathit{lift}$ hold, respectively.
Completion ($q=5$) is retained. Before completion, if maintenance holds but progression does not, the phase is unchanged.
Loss of maintenance triggers \emph{rollback} to phase 1, followed by \emph{re-advancement} using the same physical state.
Each update traverses all consecutive transitions whose conditions hold.
For example, losing the grasp during ClearPath while stable access remains yields $3\to1\to2$, so grasping must be retried.

After the second round, we freeze the thresholds and aligned waypoints and collect the final demonstrations (\cref{fig:method_overview}B, lower right; lines~\ref{ln:final_begin}--\ref{ln:final_end}).
For each initialized scene, the planner executes the selected scheme's waypoints and gripper actions while recording observations and physical states, from which phase tracking derives synchronized phase annotations that link the recorded bimanual actions to the physical progress they achieve.
The same tracking later evaluates autonomous rollouts (\cref{sec:policy_learning_evaluation}), so demonstrations and evaluation share identical physical criteria.

\subsection{Policy Learning \& Diagnostic Evaluation}
\label{sec:policy_learning_evaluation}

Using the final demonstrations $\mathcal{D}$, we perform \textbf{Joint Training} of six policies across the four held-in tasks (\cref{fig:method_overview}C, right; Algorithm~\ref{alg:policy_eval}, line~\ref{ln:train}).
They include image-based visuomotor policies, policies using 3D scene representations or object-motion modeling, and a pretrained vision-language-action model. After training, we freeze each policy and run \textbf{Closed-Loop Evaluation} on independent held-in layouts and held-out tasks (\cref{fig:method_overview}C, upper left; lines~\ref{ln:eval_begin}--\ref{ln:eval_end}). For each test instance, policies share the initial scene and robot state, then generate and execute bimanual actions from their own observations until task termination or the execution budget is reached.
Besides final success, we record mechanism-consistent behavior, arm-role allocation, and phase completion. Phase transitions follow State-Grounded Phase Tracking with the frozen thresholds, and the highest completed phase $b$ measures how far execution progresses. Unlike demonstration collection, autonomous rollouts assign roles by actual grasping, with $m^\star$ taking the arm that firmly holds the object at $2\to3$ as the grasper.

\begin{algorithm}[t]
\SetAlCapFnt{\footnotesize}\SetAlCapNameFnt{\footnotesize}
\caption{Policy Learning \& Diagnostic Evaluation}
\label{alg:policy_eval}
\footnotesize
\SetFuncSty{textrm}
\SetKwFunction{FTrack}{Track}
\KwIn{Final demonstrations $\mathcal{D}$; policies $\Pi$; test instances on held-in layouts and held-out tasks with expert references $\hat\tau$; \FTrack and frozen thresholds (Algorithm~\ref{alg:demo_collection}).}
\KwOut{Closed-loop records and KRPE score $E$ of each policy.}

\tcp{Joint Training}
\lForEach{$\pi\in\Pi$}{train $\pi$ on $\mathcal{D}$, then freeze it}\nllabel{ln:train}
\tcp{Closed-Loop Evaluation}
\ForEach{$\pi\in\Pi$ and test instance $s_0$}{\nllabel{ln:eval_begin}
    $(q,b)\gets(1,0)$\;
    \ForEach{step $t$ while $\pi$ acts from $s_0$ within budget}{
        $(q,b)\gets$ \FTrack{$t,q,b$}, $m^\star$ fixed by firm grasp at $2\to3$\;
    }
    record success, mechanism consistency, $m^\star$, and $b$\;\nllabel{ln:eval_end}
}
\tcp{Key-Interaction Pose Evaluation}
\ForEach{$\pi\in\Pi$ and reference $\hat\tau$ of an applicable test instance}{\nllabel{ln:krpe_begin}
    $W_{1^*},W_2,W_{2^*}\gets$ windows from phases and keypoints of $\hat\tau$\;
    $s\gets$ replay $\hat\tau$ until $W_{1^*}$ begins\;
    \ForEach{step $t\in W_{1^*}\cup W_2$}{
        $s\gets\operatorname{Step}(\pi,s)$\;
        \ForEach{$(r,k)\in\{(a,\mathrm{Contact}),(g,\mathrm{Grasp})\}$}{
            $(R_r,p_r)\gets T_k(s)^{-1}T_r(s)$\;
            $e_r(t)\gets\Phi\bigl(\|p_r-\hat p_r(t)\|_2,\ \angle(\hat R_r(t)^{\top}R_r)\bigr)$\;
        }
    }
    $E\gets\frac{1}{3}\bigl[\bar e_a(W_{1^*})+\bar e_a(W_2)+\bar e_g(W_{2^*})\bigr]$\;\nllabel{ln:krpe_end}
}

{\let\nl\relax
\vspace{2pt}
\noindent\hspace*{-1em}%
\begin{minipage}{\dimexpr\linewidth-2em-2pt\relax}
\footnotesize
Hats mark reference quantities; $T_r$, $T_k$ are world poses of end effector $r$ and keypoint $k$; $\angle(\cdot)$ is the rotation angle. $\Phi(p,\theta)=\tfrac{2}{3}p/p_0+\tfrac{1}{3}\theta/\theta_0$ with $p_0=3\,\mathrm{cm}$, $\theta_0=3^\circ$; $\bar e_r(W)$ averages $e_r$ over $W$.
\end{minipage}\par
}
\end{algorithm}

Complementing full-task evaluation, \textbf{Key-Interaction Pose Evaluation} measures spatial accuracy when policies create, maintain, and exploit grasp access (\cref{fig:method_overview}C, lower left; lines~\ref{ln:krpe_begin}--\ref{ln:krpe_end}).
The expert demonstration of each test instance serves as the reference $\hat\tau$ and defines three scoring windows: $W_{1^*}$ and $W_{2^*}$ cover the assisting arm creating access and the grasping arm exploiting it, starting once the expert end effector comes within 15\,cm of its keypoint in phases 1 and 2, respectively, while $W_2$ covers the assisting arm maintaining access throughout phase 2.
After $\hat\tau$ is replayed to the start of $W_{1^*}$, the policy takes over; at matching steps, rollout and reference each express the assisting and grasping end effectors, assigned by the reference roles, in their own current Contact and Grasp frames.
A policy that moves the object differently is thus judged by how well its gripper aligns with that object rather than by world poses.
\textbf{Keypoint-Relative Pose Error (KRPE)} combines the position and rotation errors between the two relative poses at each step, and $E$ averages it equally over the three windows.

\section{Experiments}
\label{sec:experiments}

\subsection{Experimental Setup}
\label{sec:exp_setup}

\begin{table*}
\centering
\caption{Policy observation, action, and execution interfaces.}
\label{tab:policy_learning}
\footnotesize
\setlength{\tabcolsep}{3pt}
\renewcommand{\arraystretch}{1.0}
\begin{tabularx}{\textwidth}{@{}
  >{\raggedright\arraybackslash}p{0.07\textwidth}
  >{\centering\arraybackslash}p{0.08\textwidth}
  >{\raggedright\arraybackslash}p{0.22\textwidth}
  >{\raggedright\arraybackslash}p{0.09\textwidth}
  >{\centering\arraybackslash}p{0.045\textwidth}
  >{\raggedright\arraybackslash}p{0.13\textwidth}
  >{\raggedright\arraybackslash}X@{}}
\toprule
\textbf{Policy} &
\textbf{Views} &
\textbf{Visual input} &
\textbf{Robot state} &
\textbf{Lang.} &
\textbf{Action target} &
\textbf{Generation / execution} \\
\midrule
ACT &
O &
RGB &
EE + G &
No &
EE + G &
Chunks; temporal ensembling \\
\addlinespace[2pt]
DP &
O,W &
RGB &
EE + G &
No &
EE + G &
Diffusion; RH \\
\addlinespace[2pt]
PerAct$^2$ &
F,O,S,W &
RGB-D voxels &
Gripper state &
No &
Discrete EE + G &
Keyframe targets; update after execution \\
\addlinespace[2pt]
3DP &
O,W &
Depth $\to$ XYZ cloud &
EE + G &
Yes &
EE + G &
Diffusion; RH \\
\addlinespace[2pt]
PPI &
F,O,S,W &
RGB-D scene/object clouds; DINOv2 features &
EE + G &
Yes &
EE + G &
Hierarchical diffusion: keyposes + trajectories; RH \\
\addlinespace[2pt]
$\pi_{0.5}$ &
F,W &
RGB &
J + G &
Yes &
$\Delta J$ + G &
VLA chunks; RH \\
\bottomrule
\end{tabularx}
\par\vspace{2pt}
\begin{minipage}{\textwidth}
\footnotesize
O/F: overhead/front; S/W: paired shoulder/wrist views.
EE: end-effector pose; J: joint positions; G: gripper state. EE targets are absolute in the world frame, $\Delta J$ is relative to the current joint state, Bimanual EE+G concatenates both arms' (position, quaternion, gripper) states/targets, while J+G concatenates their (joint positions, gripper) states/targets.
RH (Receding-Horizon): re-inference after partial execution. All policies execute through the simulator's native control interface: end-effector pose targets are handled by its motion planner, while joint-position targets are executed directly.
\end{minipage}
\end{table*}

Among the six policies, ACT~\cite{zhaolearning} and DP~\cite{chi2023diffusion} are image-based visuomotor policies; PerAct$^2$~\cite{grotz2024peract2} and 3DP~\cite{ze2024dp3} use 3D scene representations; PPI~\cite{yang2025gripper} incorporates object-motion modeling; and $\pi_{0.5}$~\cite{black2025pi05} uses a pretrained vision-language-action model.
\cref{tab:policy_learning} summarizes their observation inputs, action representations, and execution schemes.
Model-specific observations are constructed from recorded images, depth, and robot states; future EE poses or joint positions and gripper states form action sequences or key targets for supervision.
PPI additionally derives point-flow supervision from changes in object pose.
Language-conditioned policies use a shared instruction without mechanism, role, or phase hints. Each policy is jointly trained on 600 demonstrations from the four held-in tasks (150 each) and evaluated on 30 episodes per task: 120 held-in episodes on independent layouts and 120 held-out episodes on new objects.

\textbf{Success Rate (SR)} is the fraction of rollouts completing the full task. \textbf{Extrinsic Dexterity Consistency (EDC)} is the fraction of rollouts whose pre-grasp behavior follows the reference grasp-enabling mechanism, regardless of physical outcome.
\textbf{Role Agreement (RA)} is the percentage of rollouts in which the policy assigns grasping to the same arm as in the expert demonstration. \textbf{Phase Success Rates ($\mathrm{PSR}_{1\text{--}4}$)} give the fractions of all rollouts completing each phase, with $\mathrm{PSR}_4=\mathrm{SR}$. \textbf{Agreement-Conditioned Metrics} report SR and EDC separately for rollouts whose roles agree and disagree with the reference, and \textbf{Per-Assignment Metrics} report RA separately for left- and right-arm grasping references. \textbf{Keypoint-Relative Pose Error (KRPE)} measures the spatial accuracy of key interactions \cref{sec:policy_learning_evaluation}. Role evaluation covers the four tasks whose reference grasping arm varies with the layout: \textit{EdgePhone}, \textit{PickFork}, \textit{EdgeBoard}, and \textit{PickSpoon}. On \textit{PivotPhone} and \textit{PivotCalculator}, the layouts predominantly favor right-arm grasping, so both tasks are excluded from role evaluation and use right-arm references in KRPE; on \textit{PickPlate} and \textit{PickTray}, many equivalent contact and grasp sites make the roles interchangeable and the keypoint references ambiguous, so both tasks are excluded from role evaluation and KRPE. EDC and arm roles are annotated from rollout videos, whereas phase completion is computed by the state tracker.

\subsection{Mechanism Consistency and Task Success}
\label{sec:exp_mechanism}
\begin{figure}[t]
\centering
\includegraphics[width=\linewidth]{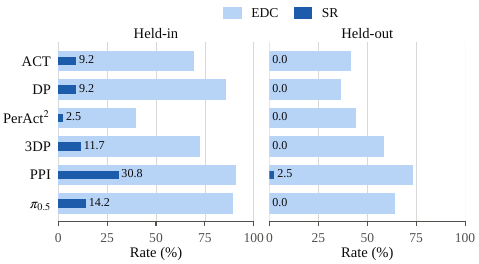}
\caption{Mechanism consistency and task success: EDC (light bars) measures how often rollouts follow the reference grasp-enabling mechanism; SR (dark bars, labeled) measures full-rollout success. Each rate averages the four tasks in its group.}
\label{fig:overview}
\end{figure}

\begin{figure*}[t]
\centering
\includegraphics[width=\textwidth]{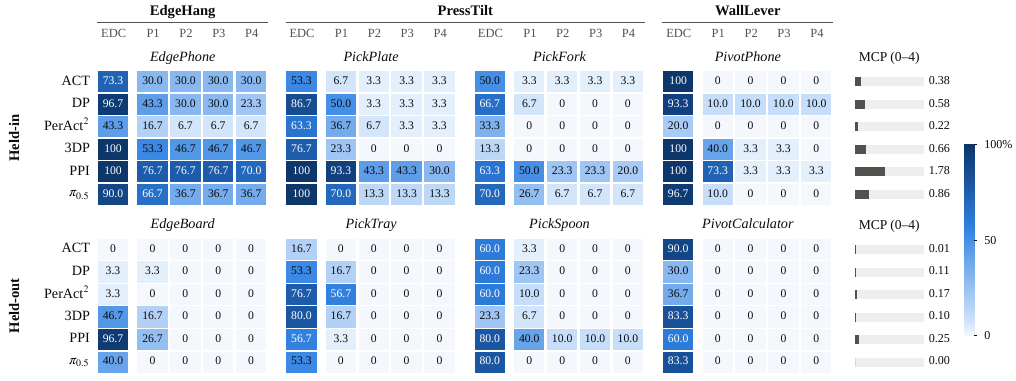}
\caption{Mechanism consistency and phase progress per task (\%). Each group lists EDC followed by $\mathrm{PSR}_1$--$\mathrm{PSR}_4$; held-out tasks appear below their held-in counterparts under the same mechanism, and the right column gives the split-average Max Completed Phase (MCP), the mean of the highest completed phase (0--4) over the four tasks for each policy and task set. Phase completion is computed by the state tracker and verified against videos.}
\label{fig:phase_heatmap}
\end{figure*}

PPI and $\pi_{0.5}$ achieve comparable held-in EDC (90.8\% and 89.2\%), yet exhibit a substantial disparity in task success, with PPI's SR exceeding that of $\pi_{0.5}$ by more than twofold (30.8\% vs.\ 14.2\%; \cref{fig:overview}).
As policies receive no mechanism prompts, these EDC values show that policies infer the appropriate mechanism from observations; the held-out evaluation below examines whether this inference is grounded in functional scene geometry or in surface appearance. PPI's superior task performance likely stems from its point-flow supervision, which explicitly models object motion responses as learning targets.
Across all policies, the majority of held-in successes (64 of 93) concentrate on \textit{EdgePhone} (\cref{fig:phase_heatmap}), where the largely self-maintaining overhang substantially reduces ongoing inter-arm coupling; in other tasks, the assisting arm must actively preserve grasp access throughout the grasping phase.
This disparity underscores that sustaining physical conditions during coordinated grasping poses a greater challenge than replicating pre-grasp motion patterns.

On held-out objects with recolored surroundings, PPI maintains 73.3\% EDC but drops to only 2.5\% SR (confined entirely to \textit{PickSpoon}), while all other policies fail completely. On \textit{EdgeBoard}, ACT, DP, and PerAct$^2$ rarely reproduce the reference behavior (EDC $\leq$ 3.3\%) and $\pi_{0.5}$ reproduces it only partially (40.0\%), whereas 3DP, which perceives color-free point clouds, and PPI, which models object-level point clouds, largely retain it (46.7\% and 96.7\%). This contrast suggests that ACT, DP, and PerAct$^2$ rely on appearance cues rather than on the support edge.
Crucially, even when the correct mechanism persists---such as for PPI on \textit{EdgeBoard} and ACT on \textit{PivotCalculator} (90.0\% EDC)---no rollout achieves task success.
This reveals that while novel object geometries may preserve macroscopic mechanism feasibility, they alter the precise contact locations, motion vectors, and bimanual synchronization required to realize it.

\subsection{Phase-Wise Execution Progress}
\label{sec:exp_phase}

Phase outcomes systematically localize where the causal chain breaks (\cref{fig:phase_heatmap}): of the 627 held-in failures, 484 never establish access, 132 establish it without securing a firm grasp, and only 11 fail thereafter (held-out: 653, 64, and 0 of 717).
On \textit{PivotPhone}, five policies exhibit at least 93.3\% EDC, yet their $\mathrm{PSR}_1$ ranges from 0\% to 73.3\%. Thus, macroscopically identical lever behaviors still differ in fine-grained attributes that EDC cannot capture, such as contact placement precision (\cref{sec:exp_krpe}).
On held-out objects, access creation degrades even under consistent mechanisms: among mechanism-consistent PPI rollouts, Phase~1 completion drops from 80.7\% to 23.9\%, suggesting that the action--effect relations fostered by point-flow supervision remain largely tied to training geometries.

Successfully established access also frequently fails to translate into a firm grasp.
For example, PPI converts all created access on \textit{EdgePhone} but less than half on \textit{PickPlate} ($\mathrm{PSR}_1$ 93.3\% vs.\ $\mathrm{PSR}_2$ 43.3\%), a contrast shared across all six policies. This confirms at the phase level the inter-arm coupling noted in \cref{sec:exp_mechanism}: access created by the assisting arm must be actively maintained while the grasping arm approaches, enters, and closes. Although \textit{PickPlate}'s multiple contact/grasp sites simplify localization, maintaining access still requires responsive closed-loop coordination between support and grasping. Among the 104 held-in rollouts that complete Phase~2, 93 achieve full success, placing dominant losses strictly before the grasp is secured.
On held-out objects, phase tracking separates an early failure in access creation, as on \textit{PivotCalculator}, from persistent breakdowns during grasp formation, as on \textit{PickTray}, where PerAct$^2$ establishes access in 56.7\% of rollouts but secures no grasp.

\subsection{Bimanual Role Coordination}
\label{sec:exp_role}
\begin{table}[t]
\centering
\caption{Arm-role agreement and its association with task success and mechanism consistency on held-in (\textit{EdgePhone}, \textit{PickFork}) and held-out (\textit{EdgeBoard}, \textit{PickSpoon}) tasks.}
\label{tab:role}
\footnotesize
\setlength{\tabcolsep}{3.2pt}
\begin{tabular}{@{}llccccc@{}}
\toprule
 & \textbf{Policy} & \textbf{RA} & RA$_{L}$/RA$_{R}$ & $n_{\text{\xmark}}$ & SR \cmark/\xmark & EDC \cmark/\xmark \\
\midrule
\multirow{7}{*}{\rotatebox{90}{Held-in}}
 & ACT & 86.7 & 83.3/90.0 & 8 & 19.2/0.0 & 67.3/25.0 \\
 & DP & 51.7 & 100.0/3.3 & 29 & 22.6/0.0 & 83.9/79.3 \\
 & PerAct$^2$ & 58.3 & 73.3/43.3 & 25 & 5.7/0.0 & 45.7/28.0 \\
 & 3DP & 60.0 & 83.3/36.7 & 24 & 38.9/0.0 & 66.7/41.7 \\
 & PPI & 60.0 & 93.3/26.7 & 24 & 66.7/12.5 & 100.0/54.2 \\
 & $\pi_{0.5}$ & 55.0 & 23.3/86.7 & 27 & 30.3/11.1 & 93.9/63.0 \\
\cmidrule(l){2-7}
 & \textit{All} & 61.9 & 76.1/47.8 & 137 & 30.0/4.4 & 75.3/52.6 \\
\midrule
\multirow{7}{*}{\rotatebox{90}{Held-out}}
 & ACT & 70.0 & 83.3/56.7 & 18 & 0.0/0.0 & 40.5/5.6 \\
 & DP & 60.0 & 83.3/36.7 & 24 & 0.0/0.0 & 33.3/29.2 \\
 & PerAct$^2$ & 45.0 & 36.7/53.3 & 33 & 0.0/0.0 & 40.7/24.2 \\
 & 3DP & 43.3 & 83.3/3.3 & 34 & 0.0/0.0 & 30.8/38.2 \\
 & PPI & 78.3 & 76.7/80.0 & 13 & 6.4/0.0 & 95.7/61.5 \\
 & $\pi_{0.5}$ & 73.3 & 46.7/100.0 & 16 & 0.0/0.0 & 75.0/18.8 \\
\cmidrule(l){2-7}
 & \textit{All} & 61.7 & 68.3/55.0 & 138 & 1.4/0.0 & 56.8/29.0 \\
\bottomrule
\end{tabular}
\par\vspace{4pt}
\parbox{\linewidth}{\footnotesize
RA$_L$/RA$_R$ measure role agreement when the reference grasping arm is left/right, respectively. \cmark/\xmark\ denote agreement/disagreement with the reference roles. SR and EDC are calculated separately for these groups.
$n_{\text{\xmark}}$ counts mismatches.
Each policy has 60 held-in and 60 held-out rollouts: 30 with a left-arm and 30 with a right-arm reference in each set.
Roles are identified from videos even when grasping fails; \textit{All} pools six policies.
All values except $n_{\text{\xmark}}$ are percentages.
}
\end{table}

Arm-role allocation is also linked to execution outcomes: on the held-in role tasks, agreement with the reference assignment coincides with higher SR (30.0\% vs.\ 4.4\%) and EDC (75.3\% vs.\ 52.6\%), a gap observed for both reference-arm groups (\cref{tab:role}). Deviations in arm roles and mechanisms may thus stem from a shared failure mode: difficulty in adapting to layout-induced kinematic constraints. Nevertheless, reference role agreement is associated with, but not strictly required for, task success. In particular, PPI and $\pi_{0.5}$ each succeed in three role-disagreeing rollouts on \textit{EdgePhone}, where the largely self-maintaining overhang tolerates alternative bimanual configurations.

Dissecting per-assignment RA further reveals that aggregate agreement often conceals pronounced single-arm biases.
Under balanced evaluation layouts, a policy that rigidly assigns grasping to the same arm trivially achieves a deceptive 50\% overall RA.
Indeed, while DP and $\pi_{0.5}$ exhibit comparable overall RA (51.7\% and 55.0\%), their per-assignment metrics expose extreme handedness: DP almost exclusively grasps with the left arm (100.0\% vs.\ 3.3\% on left/right references), whereas $\pi_{0.5}$ displays the opposite preference (23.3\% vs.\ 86.7\%). Among all policies, only ACT demonstrates genuine bidirectional adaptability (83.3\% and 90.0\%).
Under held-out generalization, while pooled RA remains nearly unchanged (61.9\% to 61.7\%), SR among agreeing rollouts collapses to 1.4\%.
Role agreement alone therefore does not ensure success on new objects, which motivates a closer examination of interaction precision.

\subsection{Key-Interaction Pose Accuracy}
\label{sec:exp_krpe}
\begin{figure*}[t]
\centering
\includegraphics[width=\textwidth]{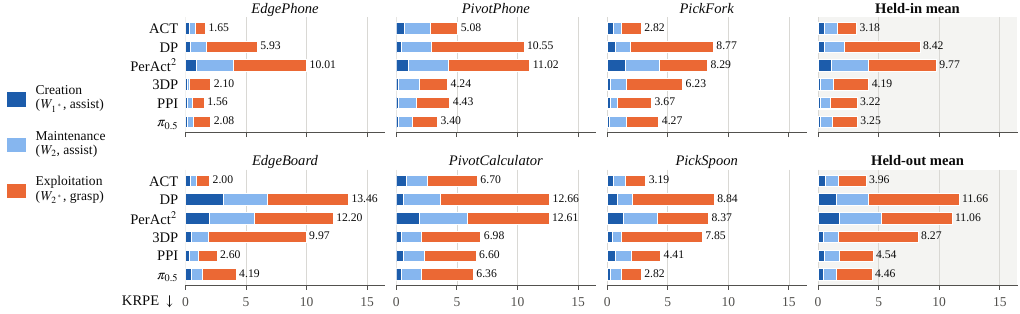}
\caption{Keypoint-relative pose error (KRPE) during critical interactions (lower is better). Bars compare policy and expert gripper poses in local keypoint frames; the rightmost columns average the three tasks in the held-in (top) and held-out (bottom) rows. Colors denote access creation (dark blue, assisting arm), maintenance (light blue, assisting arm), and exploitation (orange, grasping arm). Each segment's length shows its contribution to the total KRPE, with the three windows weighted equally.}
\label{fig:krpe}
\end{figure*}

Under expert-initialized near-contact conditions, 3DP aligns its grippers markedly better than DP, with roughly half the held-in KRPE (4.19 vs.\ 8.42; \cref{fig:krpe}). Across all policies, the exploitation window contributes the largest pose errors, consistent with the grasp-formation losses in \cref{sec:exp_phase}. On held-out objects, KRPE rises for every policy (by 13--97\%; 3DP from 4.19 to 8.27), quantifying the contact shifts noted in \cref{sec:exp_mechanism}. Crucially, accurate gripper placement does not guarantee successful manipulation. On \textit{PivotPhone}, ACT's contact error is four times that of PPI, consistent with its failure to create access; yet $\pi_{0.5}$ matches PPI's accuracy while still failing far more often.
Similarly, on \textit{PickSpoon}, ACT and $\pi_{0.5}$ achieve lower pose errors than PPI (3.19 and 2.82 vs.\ 4.41), but rarely establish access in full rollouts ($\mathrm{PSR}_1$ of 3.3\% and 0\% vs.\ 40.0\%). This gap arises because pose evaluation tests policies from near-contact setups with pre-assigned arm roles, whereas autonomous rollouts require policies to organize the full approach from the very beginning. Because matching expert poses does not ensure effective physical interaction, KRPE and phase tracking provide complementary insights: one measures geometric precision, while the other verifies physical execution.

These results specify where improvement must occur: raising $\mathrm{PSR}_1$ requires contact that shifts the object into a graspable state; converting access into a firm grasp requires the assisting arm to preserve access while the grasping arm enters and closes; and translating low KRPE into task success requires policies to organize the full approach autonomously.

\section{Conclusion}

We presented Bi-TRACE, a diagnostic benchmark that examines whether policies for bimanual occluded grasping create, maintain, and exploit the physical conditions on which later actions depend, combining layout-adaptive demonstrations, state-grounded phase tracking, and keypoint-relative interaction evaluation. Experiments with six general-purpose policies on held-in and held-out tasks reveal that failures concentrate in access creation and coordinated grasp formation, often despite mechanism-consistent behavior, and that lower pose errors under near-contact initialization may not translate into better full-task execution. Functional keypoints can guide training for precise contact and grasp alignment, and state-grounded phase supervision can help policies track execution progress and select appropriate subgoals. By actively monitoring physical states, future policies can detect prerequisite failures, roll back, and retry subgoals rather than blindly advancing, paving the way toward robust bimanual coordination and complex multi-stage manipulation.
\balance
\setstretch{0.97}
\bibliographystyle{IEEEtran}
\bibliography{reference_header,reference}

\end{document}